\pdfoutput=1

\documentclass[11pt]{article}
\usepackage{acl}
\usepackage{multirow}
\usepackage{times}
\usepackage{latexsym}

\usepackage[T1]{fontenc}

\usepackage[utf8]{inputenc}

\usepackage{microtype}

\usepackage{amsmath,amsfonts,amsthm,amssymb,mathtools,bbm,bm}
\newcommand{\real}{\mathbbm{R}}

\theoremstyle{definition}
\newtheorem{definition}{Definition}[section]

\usepackage{graphicx}
\usepackage{caption}
\usepackage{subcaption}
\usepackage{tikz}
\usetikzlibrary{arrows}

\title{Systematicity, Compositionality and Transitivity of Deep NLP Models:\\a Metamorphic Testing Perspective}

\author{Edoardo Manino \\
    University of Manchester \\
    United Kingdom \\
    \And
    Julia Rozanova \\
    University of Manchester \\
    United Kingdom \\
    \And
    Danilo Carvalho \\
    University of Manchester \\
    United Kingdom \\
    \AND
    Andr\'e Freitas \\
    University of Manchester, United Kingdom \\
    Idiap Research Institute, Switzerland \\
    \And
    Lucas Cordeiro \\
    University of Manchester \\
    United Kingdom
}

\begin{document}

\maketitle

\begin{abstract}
Metamorphic testing has recently been used to check the safety of neural NLP models. Its main advantage is that it does not rely on a ground truth to generate test cases. However, existing studies are mostly concerned with robustness-like metamorphic relations, limiting the scope of linguistic properties they can test. We propose three new classes of metamorphic relations, which address the properties of systematicity, compositionality and transitivity. Unlike robustness, our relations are defined over multiple source inputs, thus increasing the number of test cases that we can produce by a polynomial factor. With them, we test the internal consistency of state-of-the-art NLP models, and show that they do not always behave according to their expected linguistic properties. Lastly, we introduce a novel graphical notation that efficiently summarises the inner structure of metamorphic relations.
\end{abstract}

\section{Introduction}
\label{sec:intro}

Many recent advances in neural models for NLP have been driven by the ability to learn from unlabeled data~\citep{Devlin2019,Liu2019}. This approach allows for training the models on large-scale corpora without the costly process of annotating them. As a result, the accuracy and complexity of state-of-the-art neural models for NLP have increased~\citep{Brown2020}.

This trend towards unlabeled data does not have a counterpart in testing NLP models. Instead, both in-distribution testing and out-of-distribution testing~\citep{Yin2019,Teney2020} rely on comparing the model's predictions to the ground truth. Similarly, attempts at \textit{probing} the internal computation of large NLP models use supervised classifiers as a diagnostic tool \citep{Ettinger2016,Belinkov2017}. 

In general, such extreme reliance on ground-truth data limits the quantity of test cases we can produce, which is a known problem in the software testing community~\citep{Barr2015}. In this regard, a promising solution is \textit{metamorphic} testing~\citep{Chen2018}. Under this paradigm, we test the internal consistency of an NLP model by checking whether it satisfies a necessary relation of its inputs and outputs~\citep{Ribeiro2020}. Consequently, metamorphic testing relies on our ability to formally express our expectations on the behaviour of an NLP model.

Still, most of the metamorphic relations proposed in the literature target the same type of behaviour, as we show in this paper. Indeed, the majority of them are \textit{robustness} relations, which require that the output of an NLP model remains stable in the face of small input perturbations \citep{Aspillaga2020}. These perturbations may involve simple typos \cite{Belinkov2018,Gao2018,Heigold2018}, replacing individual words with a synonym \cite{Li2017,Jia2019,LaMalfa2020}, or adding irrelevant information to the input \cite{Tu2021}. Due to their simple structure, robustness-like relations have been applied to the testing of several NLP tasks, including sentiment analysis~\cite{Ribeiro2020}, machine translation \cite{Sun2018}, and question answering \cite{Chan2021}. Even testing the fairness of NLP models falls in this category~\citep{Ma2020}.

At the same time, we expect state-of-the-art NLP models to exhibit a broader range of linguistic properties than just robustness. First and foremost, NLP models should generalise \textit{systematically}, i.e.\@ their ability to understand some inputs should be intrinsically connected to their ability to understand related ones~\citep{Fodor1988}. While the exact definition of systematic behaviour varies in the literature~\citep{Hupkes2020}, a common requirement is that the model's predictions are a result of a \textit{composition} of syntactic and semantic constituents of the input~\citep{Baroni2020}. Several supervised methods to test against such requirements exist~\citep{Ettinger2016,GoodwinSystematicity}, but they all rely on comparing the model's predictions to the ground truth. Likewise, \citet{Yanaka2021transitivity} interprets systematicity as the ability to generalise over \textit{transitive} relations. Their supervised method shows that current models struggle to do so.

In this paper, we propose three new classes of metamorphic relations, which are designed to test the systematicity, compositionality and transitivity of NLP models. In true metamorphic fashion, our relations do not rely on ground-truth data and scale up the generation of test cases by a polynomial factor. For each proposed relation, we provide an illustrative experiment where we test state-of-the-art models for the expected linguistic behaviours. More in detail, our main original contributions are:
\begin{itemize}
    \item \textbf{Pairwise systematicity.} First, we propose a general class of metamorphic relations to test the systematicity of NLP models (Section \ref{sec:taxonomy_double}). The relations in this class are based on \textit{pairs} of inputs, which yields a quadratic number of test cases from a single dataset. We test the pairwise systematicity of a sentiment analysis model in Section \ref{sec:sentiment_invariance}, with positive results. Then, in Section \ref{sec:geometric_double}, we give a geometrical intuition of the constraints imposed by our relations on the model's embedding space.
    
    \item \textbf{Pairwise compositionality.} Second, we modify pairwise systematicity to test the presence of compositional constituents in the hidden layers of neural models (Section \ref{sec:compositional_relations}). Accordingly, we test the pairwise compositionality of a natural language inference (NLI) model in Section \ref{sec:compositional_entailment}, and show that it does not behave in a compositional way.
    
    \item \textbf{Three-way transitivity.} Third, we introduce a class of relations to test the internal transitivity of an NLP model (Section \ref{sec:triplet_relations}). These relations are defined over \textit{triplets} of source inputs. In Section \ref{sec:triplet_transitivity}, we test a state-of-the-art model that predicts the lexical relation of words (synonymy, hypernymy), and show that it does not behave in a transitive way.
    
    \item \textbf{Graphical notation.} Fourth, we propose a formal graphical notation for NLP metamorphic relations, that efficiently expresses their internal structure (Section \ref{sec:background}).
    
    \item \textbf{Taxonomy of existing work.} Fifth, we review the existing literature on metamorphic testing for NLP, and show that the relations proposed therein share the same structure with a single source input (Section \ref{sec:taxonomy_single}).
\end{itemize}
Lastly, in Section \ref{sec:conclusion} we conclude and outline possible future work. We discuss the ethical implications of our work in Appendix A. We provide a quick-reference guide to our contribution in Appendix B.
The code of our experiments and reproducibility checklist are available at \url{https://doi.org/10.5281/zenodo.5703459}.

\section{A graphical notation for NLP metamorphic relations}
\label{sec:background}

This section gives preliminary definitions and proposes a compact graphical notation for NLP metamorphic relations.
\begin{definition}[NLP model]
\label{def:model}
    Let $f:\mathcal{X}\to\mathcal{Y}$ be a machine learning model that maps a textual input $\mathbf{x}\in\mathcal{X}$ to a suitable output $\mathbf{Y}\in\mathcal{Y}$. Here, we assume that $f$ is a neural network, and $\mathcal{Y}\equiv\real^k$ is either a $k$-dimensional embedding space or the soft-max output of a $k$-class classifier.
\end{definition}
In general, a metamorphic relation can be defined as~\citep{Chen2018}:

\begin{definition}[Metamorphic relation]
\label{def:relation}
    A metamorphic relation $R$ is a property of $f$ across multiple inputs and outputs $(\mathbf{x}_1,\dots,\mathbf{x}_v,f(\mathbf{x}_1),\dots,f(\mathbf{x}_v))$, such that $R\subseteq \mathcal{X}_1\times\dots\times\mathcal{X}_v\times\mathcal{Y}_1\times\dots\times\mathcal{Y}_v$.
\end{definition}

However, we are interested in the internal structure of such a relation. Thus, let us discriminate between two types of inputs~\citep{Chen2018}:

\begin{definition}[Source inputs]
\label{def:source_input}
    Given a relation $R$ with $v$ inputs, let $(\mathbf{x}_1,\dots,\mathbf{x}_u)$ with $u\leq v$ be the sequence of source inputs. These can be chosen freely, e.g.\@ by extracting them from a dataset $\mathcal{D}$.
\end{definition}

\begin{definition}[Follow-up inputs]
\label{def:follow_up_input}
    Given a relation $R$ with $u$ source inputs, let $(\mathbf{x}_{u+1},\dots,\mathbf{x}_v)$ with $u\leq v$ be the sequence of follow-up inputs. These are computed by a transformation of the source inputs $\mathbf{x}_i=T_i(\mathbf{x}_1,\dots,\mathbf{x}_u)$ for $i\in[u+1,v]$.
\end{definition}

Furthermore, all the relations in this paper prescribe specific conditions over the model's output:

\begin{definition}[Output property]
\label{def:output_property}
    Define $P\subseteq\mathcal{Y}_1,\dots,\mathcal{Y}_v$ as a relation over the output. Here, we always write it in decidable first-order logic.
\end{definition}

Altogether, the structure of an NLP metamorphic relation can be easily described in graphical form. To do so, we introduce the following compact notation (see example in Figure~\ref{fig:theor_graph_single}). Textual variables are represented as circles, whereas numerical variables (e.g.\@ embeddings, softmax outputs) are squares. Moreover, source inputs are shaded in grey, while all other nodes are in white. Arrows represent the neural function $f$ and the transformation $T_i$. Lastly, the output property $P$ is linked to the relevant nodes with dashed lines.

\section{A taxonomy of existing NLP metamorphic relations}
\label{sec:taxonomy_single}

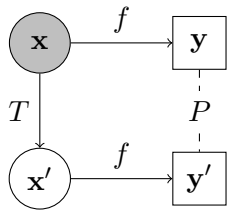
\begin{figure}[h]
\centering
    \begin{tikzpicture}[]
        \node[draw,minimum size=0.8cm,circle,fill=lightgray](a){$\mathbf{x}$};
        \node[draw,minimum size=0.8cm,circle,below of=a,node distance=1.8cm](b){$\mathbf{x}'$};
        \node[draw,minimum size=0.7cm,right of=a,node distance=2.1cm](c){$\mathbf{y}$};
        \node[draw,minimum size=0.7cm,below of=c,node distance=1.8cm](d){$\mathbf{y}'$};
        
        \path[->] (a) edge node [above] {$f$} (c);
        \path[->] (b) edge node [above] {$f$} (d);
        
        \path[->] (a) edge node [left] {$T$} (b);
        
        \draw[dashed] (c) -- (d) node [midway, fill=white] {$P$};
    \end{tikzpicture}
\caption{Structure of a single-input metamorphic relation. Property $P$ expresses how the output of model $f$ should change when the source input $\mathbf{x}$ is modified via $T$. Most relations in the literature follow this structure.}
\label{fig:theor_graph_single}
\end{figure}

Most of the existing literature on NLP metamorphic testing proposes relations that fit in the structure of Figure~\ref{fig:theor_graph_single}. Due to their reliance on just one source input, we refer to these metamorphic relations as \textit{single-input}. The individual differences among them can be ascribed to the specific transformation $T$ and property $P$. The present section derives a taxonomy of existing NLP relations by organising them along these two axes $T$ and $P$.

The transformation $T$ is defined over the input text and thus allows for considerable creative freedom. A list of common options is presented here:

\begin{itemize}
    \item \textbf{Character-level $T$.} Character-level transformations are typically used to introduce noise in the input. Examples include replacing individual characters with a neighbouring one on a computer keyboard~\citep{Belinkov2018} or a random one~\citep{Heigold2018}. More aggressive transformations may involve swapping neighbouring characters~\citep{Belinkov2018,Gao2018,Heigold2018} and shuffling a subset of the characters in a word~\citep{Belinkov2018}. Alternatively, a collection of real-world typos can be retrieved from datasets with edit history (e.g. Wikipedia)~\citep{Belinkov2018}.
    
    \item \textbf{Word-level $T$.} A common word-level transformation involves replacing words with their synonym~\citep{Li2017}. This operation has been shown to produce adversarial examples in~\citep{Jia2019,LaMalfa2020}. The use of antonyms has also been explored in~\citet{Tu2021}. In contrast, changing the gender of keywords in the input text can reveal the social biases of an NLP model~\citep{Ma2020}. Similarly, swapping keywords in the context of a question-answer (QA) system can reveal inconsistent answers~\citep{Ribeiro2020}. In the same vein, \citet{Fadaee2020} and \citet{Dankers2021} shows the volatility of neural translation models to minor word-level transformations of the input.
    
    \item \textbf{Sentence-level $T$.} Removal or concatenation of entire sentences from the input text has been tried too. \citet{Aspillaga2020} experiments with adding positive and negative tautologies at the end of the input. Similarly, \citet{Ribeiro2020} propose to concatenate both well-formed sentences and randomly-generated URLs. More generally, the whole input text can have its sentences shuffled~\citep{Tu2021} or paraphrased~\citep{Li2017}.
\end{itemize}

Regarding the output property $P$, the current literature only offers three choices. We list them here, alongside their first-order logic formulation:

\begin{itemize}
    \item \textbf{Equivalence $P$.} Robustness relations require that the output does not change in the face of small input perturbations. Thus, we need a notion of equivalence between the source output $\mathbf{y}$ and its follow-up $\mathbf{y}'$ (see Figure \ref{fig:theor_graph_single}). For classification models, we can express it via the softmax output $\mathbf{y}\!=\!(y_1,\dots,y_c)$ as:
    \begin{equation}
    \label{eq:prop_equiv}
        P_{eq}:\quad\exists i\:\forall j\!\neq\!i\:
        (y_i>y_j)\land(y_i'>y_{j}')
    \end{equation}
 \noindent where $i$ is the predicted class. In rarer cases, where the output is textual, verbatim comparison can be used~\citep{Sun2018}.
    
    \item \textbf{Similarity $P$.} For other applications, the equivalence property cannot be applied. For example, when testing QA systems, we want to detect similar but not identical answers. In such cases, we can define a similarity score $s(\mathbf{y},\mathbf{y}')\in \real$, e.g.\@ cosine similarity between the embeddings of the two answers~\citep{Tu2021}. With it, we can write similarity as:
    \begin{equation}
    \label{eq:prop_simil}
        P_{sim}:\quad s(\mathbf{y},\mathbf{y}')>\theta
    \end{equation}
\noindent where $\theta$ is an arbitrary threshold chosen according to the user's domain knowledge.
    
    \item \textbf{Order $P$.} At the same time, we can establish an order relation between the two outputs $\mathbf{y}$ and $\mathbf{y}'$. This order relation is useful in conjunction with transformations that have a monotonic effect on the output. For example, concatenating positive sentences to the input of a sentiment analysis system~\citep{Ribeiro2020}. In such cases, let us define an order score $s(\mathbf{y})\in \real$, and write the output property as:
    \begin{equation}
    \label{eq:prop_order}
        P_{ord}:\quad s(\mathbf{y})<s(\mathbf{y}')
    \end{equation}
\end{itemize}

In Sections \ref{sec:taxonomy_double}, \ref{sec:compositional_relations} and \ref{sec:triplet_relations} we employ some of the transformations $T$ and properties $P$ defined here as building blocks for new metamorphic relations.

\section{Pairwise NLP metamorphic relations for testing systematicity}
\label{sec:taxonomy_double}

\begin{figure}[h]
\centering
    \begin{tikzpicture}[]
        \node[draw,minimum size=0.8cm,circle,fill=lightgray](a){$\mathbf{x}_1$};
        \node[draw,minimum size=0.8cm,circle,below of=a,node distance=1.8cm](b){$\mathbf{x}_1'$};
        \node[draw,minimum size=0.7cm,right of=a,node distance=1.8cm](c){$\mathbf{y}_1$};
        \node[draw,minimum size=0.7cm,below of=c,node distance=1.8cm](d){$\mathbf{y}_1'$};
        
        \node[draw,minimum size=0.7cm,right of=c,node distance=1.8cm](e){$\mathbf{y}_2$};
        \node[draw,minimum size=0.7cm,below of=e,node distance=1.8cm](f){$\mathbf{y}_2'$};
        \node[draw,minimum size=0.8cm,circle,fill=lightgray,right of=e,node distance=1.8cm](g){$\mathbf{x}_2$};
        \node[draw,minimum size=0.8cm,circle,below of=g,node distance=1.8cm](h){$\mathbf{x}_2'$};
        
        \node[draw=none,below right of=c,xshift=0.2cm,yshift=-0.25cm](i){$P$};
        
        \path[->] (a) edge node [above] {$f$} (c);
        \path[->] (b) edge node [above] {$f$} (d);
        \path[->] (g) edge node [above] {$f$} (e);
        \path[->] (h) edge node [above] {$f$} (f);
        
        \path[->] (a) edge node [left] {$T$} (b);
        \path[->] (g) edge node [right] {$T$} (h);
        
        \draw[dashed] (c) -- (i);
        \draw[dashed] (d) -- (i);
        \draw[dashed] (e) -- (i);
        \draw[dashed] (f) -- (i);
    \end{tikzpicture}
\caption{Structure of pairwise-systematicity relations. The two source inputs allow us to establish a baseline for the behaviour of model $f$, and test whether it changes according to expectations once $T$ is applied.}
\label{fig:theor_graph_double}
\end{figure}
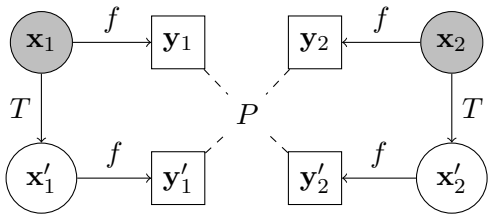

We introduce a new class of metamorphic relations to test the systematicity of NLP models. Here, we take the general definition of systematicity in \citet{Fodor1988}, which states that the predictions of an NLP model across related inputs should be intrinsically connected and express it as a metamorphic relation (see Figure \ref{fig:theor_graph_double}). Since we do not want to rely on ground-truth data, we first establish a baseline for the model's behaviour by comparing its predictions across two different source inputs. Then, we perturb both source inputs via the same transformation and test whether the model's behaviour changes accordingly.

More formally, we define \textit{pairwise-systematicity} relations as follows. Let $\mathbf{x}_1,\mathbf{x}_2\in\mathcal{D}$ be a pair of source inputs, and $\mathbf{x}_1',\mathbf{x}_2'$ their corresponding follow-up inputs via transformation $T$. Furthermore, denote with $\mathbf{y}_1,\mathbf{y}_2,\mathbf{y}_1',\mathbf{y}_2'$ the outputs produced by model $f$. Finally, define the output property $P$ in the following form:
\begin{equation}
\label{eq:systematicity_out_prop}
    P:\quad P_{src}(\mathbf{y}_1,\mathbf{y}_2)\implies P_{flw}(\mathbf{y}_1',\mathbf{y}_2')
\end{equation}

Note that this definition does not rely on ground-truth data. In fact, we trust the model's predictions $(\mathbf{y}_1,\mathbf{y}_2)$ over the source inputs to establish our premise $P_{src}$. The actual test checks whether transforming the source inputs with $T$ produces outputs that satisfy the expected property $P_{fwl}$. Any violation of this property, i.e.\@ when $P_{src}\land\neg P_{fwl}$, reveals an inconsistency in the model's predictions that breaks the user's expectation of systematic behaviour. In Section \ref{sec:geometric_double}, we give an intuitive geometrical explanation of the type of constraints imposed by pairwise-systematicity relations on the embedding space of a neural NLP model.

A hidden advantage of metamorphic relations with multiple source inputs (see also Sections \ref{sec:compositional_relations} and \ref{sec:triplet_relations}) is that they naturally produce more test cases than single-input ones. In the case of pairwise systematicity, each input in the pair $(\mathbf{x}_1,\mathbf{x}_2)$ is extracted from the same dataset $\mathcal{D}$. Thus, a dataset with $|\mathcal{D}|=k$ entries generates an $O(k^2)$ number of test cases, as opposed to $O(k)$ for single-input relations. We see an example of this in Section \ref{sec:sentiment_invariance}.

\subsection{Illustrative example: pairwise systematicity of sentiment analysis}
\label{sec:sentiment_invariance}

Now, let us apply the pairwise-systematicity relation structure shown in Figure~\ref{fig:theor_graph_double} to a sentiment analysis task. To do so, we choose the following:
\begin{itemize}
    \item \textbf{Transformation $T$.} For each source input $\mathbf{x}_i$, we create a follow-up input $\mathbf{x}_i'=T(\mathbf{x}_i)$ by concatenating a short sentence to it. A list of all transformations we use is in Table~\ref{tab:sentiment_invariance_transform}.
    
    \item \textbf{Output premise $P_{src}$.} Let $s_{pos}(\mathbf{y}_1)$ and $s_{pos}(\mathbf{y}_2)$ be the (positive) sentiment scores predicted by model $f$. Define the baseline behaviour of $f$ as the order property $P_{src}\!=\!P_{ord}$ between these two scores (see Equation~\ref{eq:prop_order}).
    
    \item \textbf{Output hypothesis $P_{flw}$.} Let $s_{pos}(\mathbf{y}_1')$ and $s_{pos}(\mathbf{y}_2')$ be the sentiment scores of the follow-up inputs. We require that their order matches the one of the source inputs. More formally: $P_{flw}\!=\!P_{ord}$ and $P_{src}\!\implies\!P_{flw}$.
\end{itemize}

Our rationale is that the sentiment of any input shifts when we concatenate additional text. If we have ground-truth information on the sentiment of the text we are adding, we can test whether our predictions shift in the expected direction. For instance, concatenating \textit{``I am very happy''} should make the score of any input more positive. This is an example of single-input relation (see Section~\ref{sec:taxonomy_single} and~\citealp{Ribeiro2020}).

\begin{table}[t]
\centering
\begin{tabular}{cp{0.5\linewidth}c}
    \hline
    \textbf{Violat.} & \textbf{Concatenated Text} & \textbf{Position} \\
    \hline
    0.100 & My friends were happy, though. & End \\
    0.090 & Anyway, the sound of the rain outside was soothing. & End \\
    0.078 & As always: popcorn and coke make everything better! & End \\
    0.068 & Thank you. & Start \\
    0.057 & I watched this movie with my brother. & Start \\
    0.045 & Here is my review: & Start \\
    \hline
\end{tabular}%
\caption{Input transformations sorted by decreasing proportion of violated test cases.}
\label{tab:sentiment_invariance_transform}
\end{table}

However, if we do not have such ground truth, we can still test our model. We do so by considering a pair of inputs $(\mathbf{x}_1,\mathbf{x}_2)$, and concatenating the same text to both of them. Then, whenever $\mathbf{x}_1$ is predicted more positive than $\mathbf{x}_2$, we require that its transformed version $\mathbf{x}_1'$ is also more positive than $\mathbf{x}_2'$ and vice versa. This is pairwise systematicity.

\paragraph{Experiment description and results.} We select a fine-tuned version of RoBERTa~\citep{Liu2019} for sentiment analysis from the HuggingFace library.\footnote{\url{https://huggingface.co/siebert/sentiment-roberta-large-english}}. We choose 10,605 movie reviews from~\citet{Socher2013} as our dataset $\mathcal{D}$. From it, we generate all $112$M+ possible source input pairs. We repeat our experiment with different neutral transformations $T$, and report their aggregated results in Table \ref{tab:sentiment_invariance_transform}. Note how the proportion of violated relations varies across different transformations. Yet, the model's behaviour is fairly systematic, never exceeding $10$\% violations.

We get a different picture by counting the number of violations per each source input $x_i\in\mathcal{D}$ (see Table \ref{tab:sentiment_invariance_inputs}). There, we can see that some inputs are more likely to make the source order $P_{src}(\mathbf{y}_1,\mathbf{y}_2)$ unstable across all the transformations $T$. Interestingly, a quick read through the reviews in Table \ref{tab:sentiment_invariance_inputs} shows that they are all misclassified. Thus, we can conclude that pairwise-systematicity testing reveals a different issue in the model $f$ than classic non-metamorphic testing. For this reason, we encourage practitioners to perform both types of testing on their NLP models, as it will give a clearer picture of their strengths and weaknesses.

\begin{table}[t]
\centering
\begin{tabular}{cp{0.55\linewidth}c}
    \hline
    \textbf{Violat.} & \textbf{Source Input} & \textbf{Pred.} \\
    \hline
    0.269 & This isn't a ``Friday'' worth waiting for. & Pos \\
    0.259 & The audience when I saw this one was chuckling at all the wrong times, and that's a bad sign when they're supposed to be having a collective heart attack. & Pos \\
    \dots & \multicolumn{1}{c}{\dots} & \dots \\
    0.000 & As a director, Paxton is surprisingly brilliant, deftly sewing together what could have been a confusing and horrifying vision into an intense and engrossing head-trip. & Neg \\
    0.000 & Intended to be a comedy about relationships, this wretched work falls flat in just about every conceivable area. & Pos \\
    \hline
\end{tabular}
\caption{Source inputs and their predicted sentiment, sorted by the number violated pairs they appear in.}
\label{tab:sentiment_invariance_inputs}
\end{table}


\subsection{Geometric interpretation of pairwise systematicity}
\label{sec:geometric_double}

Metamorphic relations impose constraints between the inputs and outputs while treating the model $f$ as a black box~\citep{Chen2018}. Still, in neural networks, it is possible to trace the effect of a relation $R$ on the hidden layers. Here, we give a geometric explanation of the type of constraints pairwise-systematicity relations put on the last embedding space of a neural NLP model.

\begin{figure*}[t]
\centering
        \includegraphics[width=0.37\textwidth]{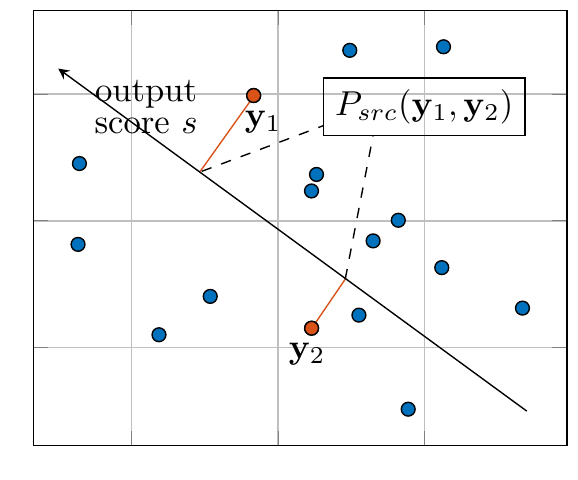}
    \hspace{0.5cm}
        \includegraphics[width=0.37\textwidth]{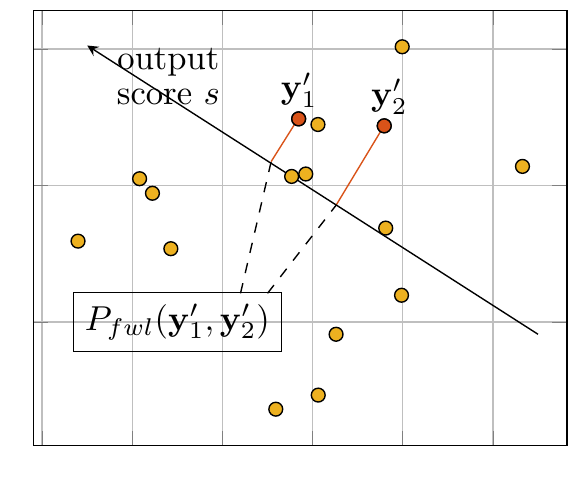}
    \hspace{0.5cm}
\caption{Pairwise systematicity relates pairs of source outputs (left) to pairs of follow-up outputs (right) in the embedding space. For the pairwise systematicity relations in Section~\ref{sec:sentiment_invariance}, the order of each pair along dimension $s$ must be preserved, as shown in this example.}
\label{fig:geom_double}
\end{figure*}

To this end, let us consider the relations in Section~\ref{sec:sentiment_invariance}. Recall, that model $f$ outputs a sentiment score $s(\mathbf{y})$, which is a one-dimensional projection of the hidden representations (see Figure~\ref{fig:geom_double}). Accordingly, the premise $P_{src}$ and hypothesis $P_{flw}$ are only concerned with the position of each representation $\mathbf{y}$ along direction $s$. However, since the source and follow-up inputs differ due to transformation $T$, the two output properties $P_{src}$ and $P_{flw}$ act on different points in the embedding space. Once we require that $P_{src}\!\implies\!P_{flw}$, we set the expectation that $f$ is exceptionally consistent at mapping pairs of inputs $(\mathbf{x}_1,\mathbf{x}_2)$ onto space $\mathcal{Y}$ in the same order.

Similar considerations apply if $P_{src}$ and $P_{flw}$ are based on equality or similarity rather than order. Indeed, equality (see Equation~\ref{eq:prop_equiv}) is defined over the softmax outputs, which are affine combinations of the embeddings~\citep{Bishop2006}. In such case, the condition $P_{src}\!\!\implies\!\!P_{flw}$ translates to a requirement that if the source inputs are both mapped to the same half-space, the follow-up inputs should be too. Conversely, similarity (Equation~\ref{eq:prop_simil}) defines a measure on the embedding space. Source inputs that are within a certain threshold $\theta$ should be matched by follow-up inputs that are also close.

Let us stress here that such geometric constraints are a direct consequence of the metamorphic relation we choose. This is a fundamentally different mechanism to the one explored by \citet{Allen2019}, where the linear relationship between the representations of related words is explained as an emergent behaviour of the probability of words occurring in similar contexts. In the following Section \ref{sec:compositional_relations}, we introduce a class of pairwise relations where the output premise and hypothesis are defined over separate embedding spaces.

\section{Pairwise NLP metamorphic relations for testing compositionality}
\label{sec:compositional_relations}

Many \textit{probing} works train simple supervised classifiers on top of the hidden representations of an NLP model (e.g.\@ \citealp{hewitt-manning}). These classifiers, called \textit{probes}, can reveal whether the neural model has learnt to recognise some fundamental constituents of the input language early on. The presence of such building blocks can be a sign that an NLP model exhibits compositional behaviour~\citep{Baroni2020}. Here, we propose to test the presence of compositional constituents in the hidden layers via metamorphic testing. To this end, we turn towards a stricter definition of mathematical compositionality of the neural network behaviour, rather than global linguistic compositionality, which is harder to define \citep{Dankers2021}.

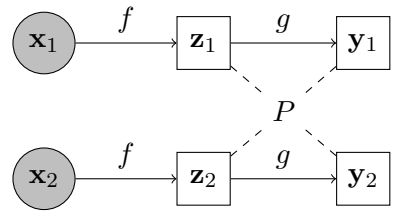
\begin{figure}[h]
\centering
    \begin{tikzpicture}[]
        \node[draw,minimum size=0.8cm,circle,fill=lightgray](a){$\mathbf{x}_1$};
        \node[draw,minimum size=0.8cm,circle,below of=a,node distance=1.8cm,fill=lightgray](b){$\mathbf{x}_2$};
        \node[draw,minimum size=0.7cm,right of=a,node distance=2.1cm](c){$\mathbf{z}_1$};
        \node[draw,minimum size=0.7cm,below of=c,node distance=1.8cm](d){$\mathbf{z}_2$};
        \node[draw,minimum size=0.7cm,right of=c,node distance=2.1cm](e){$\mathbf{y}_1$};
        \node[draw,minimum size=0.7cm,below of=e,node distance=1.8cm](f){$\mathbf{y}_2$};
        
        \path[->] (a) edge node [above] {$f$} (c);
        \path[->] (b) edge node [above] {$f$} (d);
        \path[->] (c) edge node [above] {$g$} (e);
        \path[->] (d) edge node [above] {$g$} (f);
        
        \node[draw=none,below right of=c,xshift=0.35cm,yshift=-0.2cm](i){$P$};
        
        \draw[dashed] (c) -- (i);
        \draw[dashed] (d) -- (i);
        \draw[dashed] (e) -- (i);
        \draw[dashed] (f) -- (i);
    \end{tikzpicture}
\caption{Structure of pairwise-compositionality relations. Comparing the hidden representations $\mathbf{z}_1,\mathbf{z}_2$ of the source inputs reveals whether the model $f\circ g$ uses them to produce the output in a compositional fashion.}
\label{fig:theor_graph_compositional}
\end{figure}

Consider the graph in Figure~\ref{fig:theor_graph_compositional}. There, the neural model is split into the mathematical composition of two functions $f\circ g$. More precisely, $\mathbf{z}=f(\mathbf{x})$ are the hidden representation of some hidden layer, and $\mathbf{y}=g(\mathbf{z})$ is the final output. Now, let us define the output property $P$ as follows:
\begin{equation}
\label{eq:compositionality_out_prop}
    P:\quad P_{hid}(\mathbf{z}_1,\mathbf{z}_2)\implies P_{out}(\mathbf{y}_1,\mathbf{y}_2)
\end{equation}

A relation in this form allows us to express whether specific precursor signals in $\mathbf{z}$ are expected to have a direct effect on $\mathbf{y}$. In a similar way to the relations in Section \ref{sec:taxonomy_double}, both the premise $P_{hid}$ and hypothesis $P_{out}$ are established by comparing across pairs of inputs, rather than a ground-truth. In Section~\ref{sec:compositional_entailment}, we show how our technique can reveal the presence (or absence) of compositional building blocks in an NLP model.

\subsection{Illustrative example: pairwise compositionality of NLI}
\label{sec:compositional_entailment}

Here, we apply the metamorphic relation in Figure~\ref{fig:theor_graph_compositional} to test a natural language inference (NLI) model. In general, the input $\mathbf{x}=(\mathbf{x}_a,\mathbf{x}_b)$ of an NLI model is the concatenation of two pieces of text: the premise $\mathbf{x}_a$ and the hypothesis $\mathbf{x}_b$. The model's goal is to predict whether $\mathbf{x}_b$ logically follows from $\mathbf{x}_a$, i.e.\@ their \textit{entailment}.

To test whether the model's predictions exhibit a compositional behaviour, we construct our test inputs according  to~\citet{Rozanova2021supporting}. Namely, we first choose a prototypical sentence template $C(\ell)$, which we call a \textit{context}. Each context includes a placeholder token $\ell$ that can be replaced with some \textit{insertion} text. Second, we construct each input $\mathbf{x}=(C(\ell_a),C(\ell_b))$ by copying the same context twice with different insertions.

Finally, we choose the contexts $C_i$ and insertion pairs $(\ell_a,\ell_b)_j$ in such a way that their composition $(C(\ell_a),C(\ell_b))_{ij}$ has a well-definite entailment relation. Namely, the insertion pairs (see Table~\ref{tab:entailment_insertion_sorted}) are either hypernyms ($\supseteq$), hyponyms ($\subseteq$), or unrelated (none). Similarly, the contexts (see Table~\ref{tab:entailment_context_sorted}) are either \textit{upward monotone} if they preserve the insertion relation, or \textit{downward monotone} if they invert it. As a result, only the compositions $\text{Up}(\subseteq)$ and $\text{Down}(\supseteq)$ are entailed, while the rest are not.

Now, assume that both input pairs $\mathbf{x}_1=(C(\ell_a),C(\ell_b))_{i1}$ and $\mathbf{x}_2=(C(\ell_a),C(\ell_b))_{i2}$ in Figure~\ref{fig:theor_graph_compositional} are based on the same context $C_i$. We can test whether the NLI model builds its output by reasoning over the monotonicity of $C_i$ and the lexical relation of $(\ell_a,\ell_b)_j$ as follows:
\begin{itemize}
    \item \textbf{Hidden premise $P_{hid}$.} Let $\mathbf{z}$ be the embeddings of the second to last layer, for the tokens corresponding to the insertions $\ell_a$ and $\ell_b$. Train a linear probe $s_{hyp}$ on $\mathbf{z}$~\citep{Liu2019probing} to predict whether $\ell_a$ is a hypernym of $\ell_b$. Define $P_{hid}\!=\!P_{ord}$ as the order property (see Equation~\ref{eq:prop_order}) over the hypernymy scores $s_{hyp}(\mathbf{z}_1)$ and $s_{hyp}(\mathbf{z}_2)$ of the two inputs.
    
    \item \textbf{Output hypothesis $P_{out}$.} Let $s_{ent}(\mathbf{y})$ be the entailment score produced by the full neural model $f\circ g$. Moreover, define $P_{out}\!=\!P_{ord}$ as the order of the two output scores $s_{ent}(\mathbf{y}_1)$ and $s_{ent}(\mathbf{y}_2)$. Then, consider the monotonicity of the input context. If $C_i$ is downward monotone, let $P_{hid}\!\iff\!P_{out}$, since more hypernymy means more entailment. If $C_i$ is upward monotone, let $P_{hid}\!\iff\!\neg P_{out}$, since more hypernymy means less entailment.
\end{itemize}

If the NLI model $f\circ g$ had a compositional behaviour, the order $P_{hid}$ of the hypernymy scores in the hidden layer should be reflected in the order $P_{out}$ of the entailment scores in the output. Here, we show that this is not the case for a popular state-of-the-art NLI model.

\begin{table}[t]
    \centering
    \begin{tabular}{cp{0.59\linewidth}c}
        \hline
        \textbf{Violat.} & \textbf{Context} & \textbf{Mon.}\\
        \hline
        0.613 & So there is no dedicated $\langle x\rangle$ for every entity and no distinction between entity mentions and non-mention words. & Down \\
        \dots & \multicolumn{1}{c}{\dots} & \dots\\
        0.374 & There was no $\langle x\rangle$. & Down \\
        0.373 & We stood on the brink of a $\langle x\rangle$. & Up \\
        \dots & \multicolumn{1}{c}{\dots} & \dots\\
        0.254 & There are some old houses in this $\langle x\rangle$. & Up \\
        0.246 & Some $\langle x\rangle$ bloom in spring and others in autumn. & Up \\
        \hline
    \end{tabular}
    \caption{Contexts sorted by decreasing proportion of violated test cases.}
    \label{tab:entailment_context_sorted}
\end{table}

\paragraph{Experiment description and results.} We build a dataset $\mathcal{D}$ of $292$ insertions pairs and repeat our experiment with $211$ contexts, for a total of about $9$M test cases. We chose a fine-tuned version of RoBERTa for NLI as our model.\footnote{\url{https://huggingface.co/roberta-large-mnli}} The accuracy of the hypernymy probe is $0.9881$. We report the aggregated result by context in Table \ref{tab:entailment_context_sorted}. Note how downward monotone contexts lead to less compositional behaviour: overall, we have a $0.312$ proportion of violated test cases with upward contexts and $0.519$ with downward ones. This phenomenon is known in the literature~\citep{Yanaka2019}, but we show that metamorphic testing can independently detect it. If we aggregate the results by insertion pair (see Table \ref{tab:entailment_insertion_sorted}), the picture does not change. The overall proportion of violations is $0.406$, which is barely below random chance. Any deviations from this baseline can be interpreted as noise.

\begin{table}[t]
    \centering
    \begin{tabular}{ccc}
        \hline
        \textbf{Violat.} & \textbf{Insertion Pair} & \textbf{Lex. Rel.} \\
        \hline
        0.583 & (gun,woman) & none \\
        0.525 & (woman,gun) & none \\
        0.492 & (tree,cherry tree) & $\supseteq$ \\
        \dots & \dots & \dots \\
        0.410 & (fruit,apple) & $\supseteq$ \\
        0.409 & (pine,tree) & $\subseteq$ \\
        \dots & \dots & \dots \\
        0.304 & (potatoes,animals) & none \\
        0.274 & (animals,potatoes) & none \\
        \hline
    \end{tabular}
    \caption{Insertions sorted by decreasing proportion of violated test cases.}
    \label{tab:entailment_insertion_sorted}
\end{table}


\section{Three-way NLP metamorphic relations for testing transitivity}
\label{sec:triplet_relations}

An NLP model that generalises correctly should exhibit \textit{transitive} behaviour under the right circumstances~\citep{Yanaka2021transitivity}. That is, if the model predicts a transitive linguistic property over the input pairs $(\mathbf{x}_1,\mathbf{x}_2)$ and $(\mathbf{x}_2,\mathbf{x}_3)$, then it should also predict it for the pair $(\mathbf{x}_1,\mathbf{x}_3)$. Here, we propose to test this behaviour in a metamorphic way.

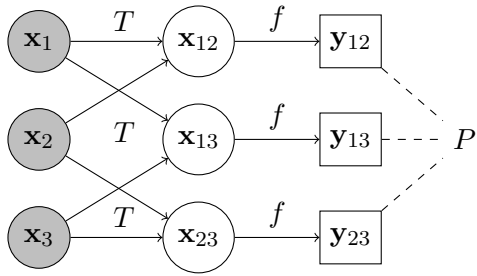
\begin{figure}[h]
\centering
    \begin{tikzpicture}[]
        \node[draw,minimum size=0.8cm,circle,fill=lightgray](a){$\mathbf{x}_1$};
        \node[draw,minimum size=0.8cm,circle,below of=a,node distance=1.3cm,fill=lightgray](b){$\mathbf{x}_2$};
        \node[draw,minimum size=0.8cm,circle,below of=b,node distance=1.3cm,fill=lightgray](c){$\mathbf{x}_3$};
        
        \node[draw,minimum size=0.8cm,circle,right of=a,node distance=2.1cm](d){$\mathbf{x}_{12}$};
        \node[draw,minimum size=0.8cm,circle,below of=d,node distance=1.3cm](e){$\mathbf{x}_{13}$};
        \node[draw,minimum size=0.8cm,circle,below of=e,node distance=1.3cm](f){$\mathbf{x}_{23}$};
        
        \node[draw,minimum size=0.7cm,right of=d,node distance=2cm](g){$\mathbf{y}_{12}$};
        \node[draw,minimum size=0.7cm,below of=g,node distance=1.3cm](h){$\mathbf{y}_{13}$};
        \node[draw,minimum size=0.7cm,below of=h,node distance=1.3cm](i){$\mathbf{y}_{23}$};
        
        \path[->] (a) edge node [above,xshift=0.1cm] {$T$} (d);
        \draw[->] (b) -- (d);
        \path[->] (a) edge node [below,xshift=0.1cm,yshift=-0.3cm] {$T$} (e);
        \draw[->] (c) -- (e);
        \path[->] (c) edge node [above,xshift=0.1cm] {$T$} (f);
        \draw[->] (b) -- (f);
        
        \path[->] (d) edge node [above] {$f$} (g);
        \path[->] (e) edge node [above] {$f$} (h);
        \path[->] (f) edge node [above] {$f$} (i);
        
        \node[draw=none,right of=h,node distance=1.5cm](j){$P$};
        
        \draw[dashed] (g) -- (j);
        \draw[dashed] (h) -- (j);
        \draw[dashed] (i) -- (j);
    \end{tikzpicture}
\caption{Structure of three-way transitivity relations. The three source inputs $\mathbf{x}_1,\mathbf{x}_2,\mathbf{x}_3$ are combined into all possible pairs. If two pairs are predicted as true by model $f$, the third must be predicted true as well.}
\label{fig:theor_graph_triplet}
\end{figure}

More specifically, let us introduce the \textit{three-way transitivity} relation in Figure \ref{fig:theor_graph_triplet}. There, the three source inputs $\mathbf{x}_1,\mathbf{x}_2,\mathbf{x}_3$ are combined to form all possible input pairs $\mathbf{x}_{ij}=(\mathbf{x}_i,\mathbf{x}_j)$. Then, we can test whether their corresponding outputs are transitive with the following output property:
\begin{equation}
\label{eq:transitivity}
P: \quad v(y_{12}) \land v(y_{23}) \Rightarrow v(y_{13})
\end{equation}
where $v(\cdot):\mathcal{Y}\to\{0,1\}$ is the Boolean prediction of model $f$. Note that the output property $P$, being defined over three outputs, has a different structure from those in Sections \ref{sec:taxonomy_single}, \ref{sec:taxonomy_double} and \ref{sec:compositional_relations}.

\subsection{Illustrative example: three-way transitivity of lexical relations}
\label{sec:triplet_transitivity}

In this section, we apply the metamorphic structure from Figure \ref{fig:theor_graph_triplet} to test the transitivity of \textit{lexical semantic relations}, e.g.\@ synonymy and hypernymy~\citep{santus-etal-2016-cogalex}. In general, learning these linguistic properties is crucial for solving several NLI tasks~\citep{Glockner2018}. Thus, we can expect an NLP model to generalise over them in a transitive way. We can test whether this is true in the following way:
\begin{itemize}
    \item \textbf{Transformation $T$.} The model $f$ we test already accepts a pair of words $\mathbf{x}_{ij}=(\mathbf{x}_i,\mathbf{x}_j)$ as input. Thus, $T$ is merely a formalism here.
    
    \item \textbf{Output Property $P$.} Property $P$ in Equation \ref{eq:transitivity} depends on the definition of $v(\cdot)$. Here, we train two classification heads on top of a pre-trained model $f$. The first $v_{syn}(\cdot)$ predicts synonymy, the second $v_{hyp}(\cdot)$ hypernymy.
\end{itemize}

Note that transitivity can be tested in a supervised fashion by comparing the model's predictions to a ground truth~\citep{Yanaka2021transitivity}. In contrast, the three-way transitivity relations we propose test the \textit{internal} transitivity of a model trained to predict lexical relations.

\paragraph{Experiment description and results.} We reproduce a state-of-the-art model for lexical relations~\citep{Wachowiak2020}, which is a fine-tuned version of the multi-lingual transformer model \texttt{xlmroberta}~\cite{conneau-etal-2020-unsupervised}. We extract the multi-lingual test set from the CogALex\_VI shared task~\cite{santus-etal-2016-cogalex}, and generate a random sample of source triplets from its corpus of words, keeping those that satisfy $v(y_{12}) \land v(y_{23})$. We present our empirical results in Table \ref{tab:transitivity_test_results}, organised by the language of the source words and lexical relation $v$ predicted by the model. As the table shows, this state-of-the-art NLP model fails to predict $v(y_{13})$ in a transitive way across all languages. This is in contrast with the results of classic supervised testing in~\citet{Wachowiak2020}, which show that their model can predict the correct lexical relations (synonym, hypernym, antonym or random) with at least 0.5 of accuracy.


\begin{table}[t]
    \centering
    \begin{tabular}{ccc}
        \hline
        \textbf{Language} & \textbf{Syn. Violat.} & \textbf{Hyp. Violat.} \\
        \hline
        English & 0.809 & 0.723 \\ 
        German & 0.760 & 0.713 \\ 
        Chinese & 0.610 & 0.606 \\ 
        Italian & 0.659 & 0.741 \\ 
        \hline
    \end{tabular}
    \caption{Proportion of violated three-way transitivity tests for a state-of-the-art lexical relation model.}
    \label{tab:transitivity_test_results}
\end{table}

\section{Conclusions and future work}
\label{sec:conclusion}

In this paper, we presented three new classes on metamorphic relations. Thanks to them, we could test the systematicity, compositionality and transitivity of state-of-the-art NLP models. The advantage of our approach is that it does not rely on ground-truth annotations. It can generate a polynomially larger number of test cases than supervised testing, revealing whether the NLP model under test is internally consistent.

Still, testing is only one side of the coin. Like in recent work about robustness \citep{Aspillaga2020}, the tested models have not been trained on a metamorphic objective (e.g.\@ as an additional loss term). We believe that doing so could improve the safety and consistency of a model's predictions.

\section*{Acknowledgements}

The work is funded by the EPSRC grant EP/T026995/1 entitled ``EnnCore: End-to-End Conceptual Guarding of Neural Architectures'' under \textit{Security for all in an AI enabled society}.

\bibliography{references}
\bibliographystyle{acl_natbib}

\appendix

\section*{Appendix A. Ethics statement}
\label{sec:ethics}

Intelligent systems are becoming increasingly widespread, and NLP models are often used as important components in their architecture. However, once these systems are deployed in the real world, there is a risk of them exhibiting biased, erratic or dangerous behaviour. In order to prevent such events from happening, it is crucial to perform a thorough testing and validation process. Indeed, this is one of the tenets of the ACM Code of Ethics and Professional Conduct\footnote{\url{https://www.acm.org/code-of-ethics}}. Namely, paragraph 2.5 therein recites \textit{``Extraordinary care should be taken to identify and mitigate potential risks in machine learning systems.''} The contributions we propose in the present paper are directed towards this goal. More specifically, we believe that metamorphic testing is a valuable tool in the model tester's arsenal, and our contributions widen its scope of application. As a result, more instances of unwanted behaviour can be identified and addressed before their impact is felt by the end user.

\section*{Appendix B. Quick-reference guide}
\label{sec:guide}

In this paper, we discuss and compare four classes of metamorphic relations. For ease of reference, we summarise them in Tables \ref{tab:guide_robustness}, \ref{tab:guide_systematicity}, \ref{tab:guide_compositionality} and \ref{tab:guide_transitivity}. These tables contain the formal definitions of the transformation $T$ and output property $P$, a concrete example of possible inputs, and a reference to the corresponding sections in the present paper.

\begin{table*}[htb]
    \centering
    \begin{tabular}{cl}
        \hline
        \multicolumn{2}{c}{\textbf{Single-input metamorphic relations}} \\ 
        \hline
        \multirow{2}{*}{Input:} & $\mathbf{x}\:=$ \fcolorbox{black}{lightgray}{The cat sat on the mat.} \\
         & $\mathbf{x}'=$ \fcolorbox{black}{white}{The pet stood onto the mat.} \\
        $T$: & \textit{replace any word of the input with a synonym.} \\
        $P$: & \multicolumn{1}{c}{$\mathbf{y}=f(\mathbf{x})\land\exists i\,\forall j\!\neq\!i\,(y_i>y_j)\land(y_i'>y_j')$} \\
        \hline
    \end{tabular}
    \captionsetup{width=.6\textwidth}
    \caption{Example of robustness relations from the literature \citep{Li2017}. Robustness relations belong to the class of single-input relations (see Section \ref{sec:taxonomy_single}).}
    \label{tab:guide_robustness}
\end{table*}

\begin{table*}[htb]
    \centering
    \begin{tabular}{cl}
        \hline
        \multicolumn{2}{c}{\textbf{Pairwise systematicity metamorphic relations}} \\
        \hline
        \multirow{4}{*}{Input:} & $\mathbf{x}_1=$ \fcolorbox{black}{lightgray}{Light, cute and forgettable.} \\
         & $\mathbf{x}_2=$ \fcolorbox{black}{lightgray}{A masterpiece four years in the making.} \\
         & $\mathbf{x}_1'=$ \fcolorbox{black}{white}{Thank you.} \fcolorbox{black}{lightgray}{Light, cute and forgettable.} \\
         & $\mathbf{x}_2'=$ \fcolorbox{black}{white}{Thank you.} \fcolorbox{black}{lightgray}{A masterpiece four years in the making.} \\
        $T$: & \textit{concatenate the text} \fcolorbox{black}{white}{Thank you.} \textit{at the beginning of the input.} \\
        $P$: & \multicolumn{1}{c}{$s_{pos}\big(f(\mathbf{x}_1)\big)>s_{pos}\big(f(\mathbf{x}_2)\big)\iff s_{pos}\big(f(\mathbf{x}_1')\big)>s_{pos}\big(f(\mathbf{x}_2')\big)$} \\
        \hline
    \end{tabular}
    \captionsetup{width=.75\textwidth}
    \caption{Example of pairwise systematicity relations defined on a sentiment analysis task (see Section \ref{sec:sentiment_invariance}).}
    \label{tab:guide_systematicity}
\end{table*}

\begin{table*}[htb]
    \centering
    \begin{tabular}{cl}
        \hline
        \multicolumn{2}{c}{\textbf{Pairwise compositionality metamorphic relations}} \\
        \hline
        \multirow{2}{*}{Input:} & $\mathbf{x}_1=$ \fcolorbox{black}{white}{There was no} \fcolorbox{black}{lightgray}{tree.} \fcolorbox{black}{white}{There was no} \fcolorbox{black}{lightgray}{cherry tree.} \\
         & $\mathbf{x}_2=$ \fcolorbox{black}{white}{There was no} \fcolorbox{black}{lightgray}{fruit.} \fcolorbox{black}{white}{There was no} \fcolorbox{black}{lightgray}{apple.} \\
        \multirow{2}{*}{Hidden:} & $f(\mathbf{x}_1)=$ \textit{contextual embeddings of the tokens (} \fcolorbox{black}{lightgray}{tree.} \fcolorbox{black}{lightgray}{cherry tree.} \textit{)}\\
         & $f(\mathbf{x}_2)=$ \textit{contextual embeddings of the tokens (} \fcolorbox{black}{lightgray}{fruit.} \fcolorbox{black}{lightgray}{apple.} \textit{)}\\
        $P$: & \multicolumn{1}{c}{$s_{hyp}\big(f(\mathbf{x}_1)\big)>s_{hyp}\big(f(\mathbf{x}_2)\big)\iff s_{ent}\big(g(f(\mathbf{x}_1))\big)>s_{ent}\big(g(f(\mathbf{x}_2))\big)$} \\
        \hline
    \end{tabular}
    \captionsetup{width=.85\textwidth}
    \caption{Example of pairwise compositionality relations defined on a natural language inference task (see Section \ref{sec:compositional_entailment}). Pairwise compositionality relations do not have a transformation $T$.}
    \label{tab:guide_compositionality}
\end{table*}

\begin{table*}[htb]
    \centering
    \begin{tabular}{cl}
        \hline
        \multicolumn{2}{c}{\textbf{Three-way transitivity metamorphic relations}} \\
        \hline
        \multirow{4}{*}{Input:} & $\mathbf{x}_1,\mathbf{x}_2,\mathbf{x}_3=$ \fcolorbox{black}{lightgray}{arrangement} \fcolorbox{black}{lightgray}{symmetrical} \fcolorbox{black}{lightgray}{together} \\
         & $\mathbf{x}_{12}=$ \textit{(} \fcolorbox{black}{white}{arrangement} \fcolorbox{black}{white}{symmetrical} \textit{)} \\
         & $\mathbf{x}_{23}=$ \textit{(} \fcolorbox{black}{white}{symmetrical} \fcolorbox{black}{white}{together} \textit{)} \\
         & $\mathbf{x}_{13}=$ \textit{(} \fcolorbox{black}{white}{arrangement} \fcolorbox{black}{white}{together} \textit{)} \\
        $T$: & \textit{choose two words from the source triplet} $\mathbf{x}_1,\mathbf{x}_2,\mathbf{x}_3$ \\
        $P_{syn}$: & \multicolumn{1}{c}{$v_{syn}\big(f(\mathbf{x}_{12})\big)\land v_{syn}\big(f(\mathbf{x}_{23})\big)\implies v_{syn}\big(f(\mathbf{x}_{13})\big)$} \\
        $P_{hyp}$: & \multicolumn{1}{c}{$v_{hyp}\big(f(\mathbf{x}_{12})\big)\land v_{hyp}\big(f(\mathbf{x}_{23})\big)\implies v_{hyp}\big(f(\mathbf{x}_{13})\big)$} \\
        \hline
    \end{tabular}
    \captionsetup{width=.65\textwidth}
    \caption{Example of three-way transitivity relations defined on the lexical relations of synonymy and hypernymy (see Section \ref{sec:triplet_transitivity}).}
    \label{tab:guide_transitivity}
\end{table*}

\end{document}